\documentclass{article}
\usepackage[preprint,nonatbib]{neurips_2021}
\usepackage{amsmath,amssymb,amsthm,mathtools}
\usepackage{graphicx,booktabs,multirow}
\usepackage[table]{xcolor}
\usepackage{natbib}
\usepackage[colorlinks=true,allcolors=blue]{hyperref}
\usepackage{enumitem}
\usepackage{pifont}

\newcommand{\signalname}{Flow Complexity}
\newcommand{\signalshort}{FC}
\newcommand{\methodname}{Flow Complexity Minimization}
\newcommand{\methodshort}{FCM}

\newcommand{\vt}{v_\theta}

\newcommand{\FC}{\mathrm{FC}}
\newcommand{\dt}{\Delta t}

\theoremstyle{plain}

\newtheorem{proposition}{Proposition}

\theoremstyle{definition}
\newtheorem{definition}{Definition}
\theoremstyle{remark}

\title{Spatial Transport of Integration Error in Generative ODEs}
\author{Songheng Yin \\ Columbia University \\ \texttt{songheng.yin@columbia.edu} }

\begin{document}
\maketitle

\begin{abstract}
A trained flow or diffusion model is usually run with only a handful of solver steps, and the
integration error this leaves behind is unevenly distributed across the image. We ask where that error
is injected and how it reaches the endpoint, and answer with a signed source-and-transport accounting
of few-step integration error, tested to first order. A perturbation experiment on five models at
$256^2$ resolution shows the learned dynamics spread local disturbances widely: near the start of
sampling, under $10\%$ of the summed endpoint response remains at the source. Signed one-step
truncation residuals, propagated through the model's own linearized dynamics, reconstruct much of the
endpoint error's direction and regional structure (cosine $0.81$--$0.87$), and a region's error owes
more to what arrives from elsewhere than to its own injection. Structure-destroying nulls, with
protocols frozen before evaluation, locate what carries the account: randomizing contribution signs
halves it, and reassigning which region receives each contribution, with content, norms, and signs
intact, destroys it entirely. Where the injections land is readable from the model itself. The
variation of its velocity or prediction field along the trajectory, a structure that emerges during
training, predicts the final per-region gap (within-image $\rho$ of $0.57$--$0.70$ on fine
trajectories, weaker from the cheap solve alone). The prediction is partial because endpoint error
depends not only on injected magnitude but on its sign, timing, and transport through the learned
dynamics. A training penalty on the injected variation lowers few-step error, so the structure is one
a model can be trained to change.
\end{abstract}

\section{Introduction}
\label{sec:intro}

A trained flow or diffusion model is usually deployed with a handful of solver steps, and the price is
numerical. Integrating the probability-flow ODE coarsely injects error that degrades the sample
\citep{ho2020denoising,song2021score,lipman2023flow,albergo2023building}. Two families of methods
reduce it. One straightens trajectories during training so that coarse integration hurts less, through
reflow \citep{liu2023flow}, consistency objectives \citep{song2023consistency,yang2024consistency}, or,
concurrent with this work, an acceleration penalty \citep{isofm2026}. The other spends inference compute
unevenly, choosing where to refine from signals read off the current sampling state
\citep{liu2025region,sdit2026semantic}. Both leave a spatially resolved numerical question largely
unmeasured. Where is coarse-solver error injected within a sample, and how do the learned dynamics
redistribute it before the endpoint? Classical numerical analysis answers in the abstract: global error
is local truncation carried through the state-transition operator \citep{hairer1993solving}. What does
not exist is a spatially resolved, empirically closed account of that propagation in trained generative
models.

\begin{figure}[tbp]
\centering
\includegraphics[width=0.65\linewidth]{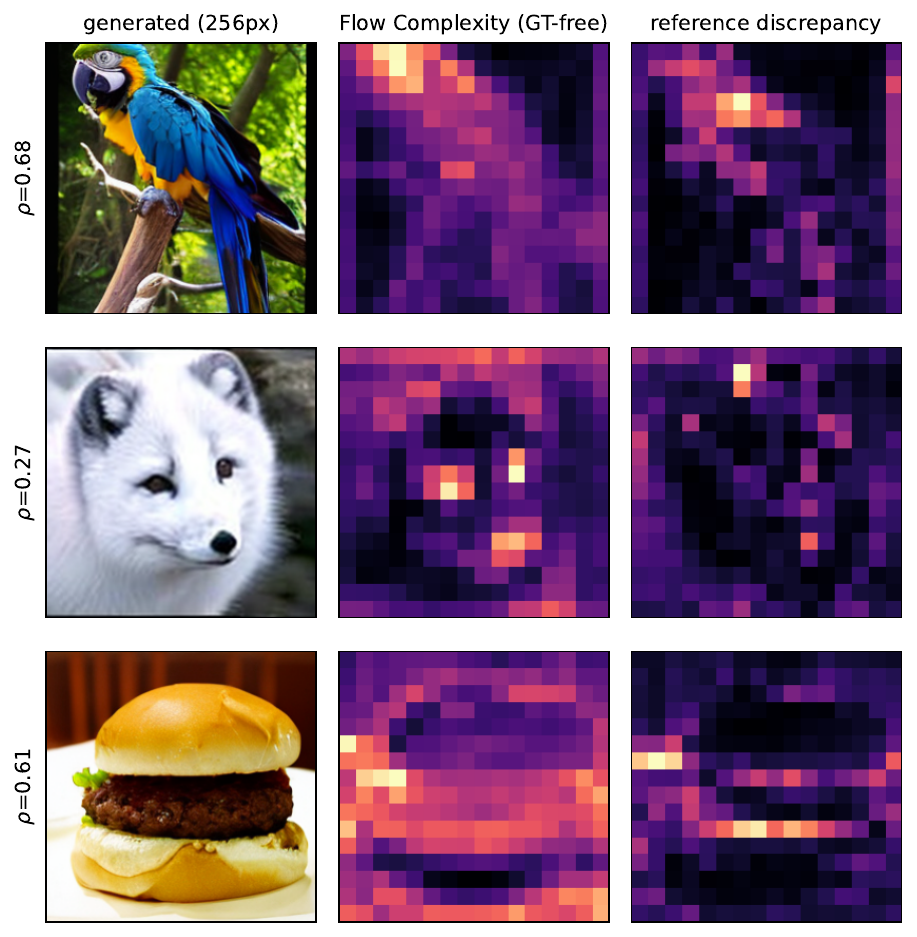}
\caption{\textbf{The spatial integrability landscape.} For SiT-XL/2 ImageNet-256 samples: the
generated image, the model-intrinsic \signalname{} map (computed from the model's own velocities,
without seeing the final error), and the per-region reference-solver discrepancy (cheap vs.\ near-exact solve).
\signalshort{} concentrates where the cheap solver actually errs (within-image Spearman $\rho$ per
row), on real $16\times16$ regions. Both are spatially structured and unevenly concentrated, not uniform.}
\label{fig:landscape}
\end{figure}

We provide one. The per-region gap between a coarse solve and an accurate reference is unevenly
concentrated across regions and repeats across sampling budgets. It appears only after training, uninformative at
initialization (Fig.~\ref{fig:landscape}). A per-region velocity statistic we call \signalname{}
predicts it, at within-image $\rho$ of $0.57$ to $0.70$ across five models, with the velocities read
along a finely integrated trajectory. Read from the first evaluations of the cheap solve itself, it
stays predictive but weaker ($\rho$ of $0.32$--$0.46$ across held-out tests on two models). The gap is
numerical rather than perceptual. Correcting where the solver errs most is not what most improves FID, so
we read \signalname{} as a diagnostic of numerical integrability, not an allocation rule for sample
quality (\S\ref{sec:notlandscape}). That prediction is only partial, and the
reason is the substance of the paper. On the two models where we can carry out the full reconstruction, each
region's endpoint error, rebuilt from the signed per-step truncation residuals propagated through the
linearized dynamics, is better explained by what reaches the region through those dynamics than by its own
injected residual. Error is injected locally, where \signalname{} runs high, then carried across the image
as the model integrates. Regularizing \signalname{} during training lowers both the injected error and
the coarse-solve error, so the structure is one a model can be trained to change.

\paragraph{Contributions.}
\begin{enumerate}[leftmargin=1.6em,itemsep=2pt,topsep=3pt]
\item \textbf{A spatial difficulty landscape (\S\ref{sec:fc},~\S\ref{sec:landscape}).} We define the
  per-region \emph{reference-solver discrepancy} of few-step sampling (cheap vs.\ near-exact solve) and
  show it is not noise but a stable, learned structure: \textbf{concentrated} (the top decile of regions
  carries $24$--$33\%$ of the error), \textbf{stable} across sampling budgets and, in controlled models,
  retraining, and \textbf{learned} (it emerges during optimization). We establish this on five models at
  $256^2$ resolution. This complements region-adaptive methods, whose reactive allocation signals are
  never checked against the reference-solver discrepancy they aim to reduce.
\item \textbf{A local diagnostic that predicts it (\S\ref{sec:fc},~\S\ref{sec:horizon}).} \signalname{}
  (\signalshort{}) is not a new dynamical quantity. For flow models it is the per-region total variation
  of the learned velocity, which a leading-order bound ties to the \emph{magnitude} of the local
  truncation the Euler residual injects (Prop.~\ref{prop:bound}). For diffusion models we read the
  matching variation of the network's prediction along the sampler trajectory, tied to discretization
  error empirically. Along fine reference trajectories it predicts final
  per-region discrepancy at within-image $\rho \approx 0.57$--$0.70$ on all five models and, at a
  \emph{matched} observation window, more accurately than the frequency, edge, attention, RAS, and
  SDiT signals region-adaptive samplers rely on (implemented as specified, not in spirit). A prediction-horizon analysis locates \emph{when}
  difficulty becomes readable (model-dependent; on the flow model already after ${\sim}10\%$ of sampling,
  $\rho \approx 0.44$).
\item \textbf{A quantitative spatial accounting of endpoint error (\S\ref{sec:transport}--\S\ref{sec:results}).}
  A perturbation protocol probes how the state-transition operator $\Phi$ of Prop.~\ref{prop:bound} moves
  injected error: near the noise end only $3$--$8\%$ of a perturbation's summed regional response stays in
  its source region, on a flow transformer, a pixel-space diffusion U-Net, and our own model, surviving
  solver refinement. On our own CelebA and ImageNet-256 models we rebuild each region's endpoint error
  from the signed per-step truncation residuals propagated through the linearized dynamics. A region's own
  injection explains only part of its final error ($\rho \approx 0.3$--$0.4$) and what arrives from
  elsewhere is more strongly associated with it (partial $\rho \approx 0.66$--$0.69$). Sign randomization
  drops the reconstruction to $\rho\approx0.3$, and destination permutation, keeping content, norms, and
  signs, destroys it outright, so the correctly targeted signed superposition carries the accounting.
  This is why \signalshort{}, a local quantity, predicts final error only in part.
  Regularizing \signalshort{} during training lowers both the injected error and the coarse-solve error, so the
  landscape is one training can modify (closely related to concurrent work,
  \citealp{isofm2026}). We do not offer it as a sampler, and we evaluate its compute-adjusted effect on
  sample quality separately (\S\ref{sec:results}).
\end{enumerate}

\section{Related work}
\label{sec:related}

\paragraph{Training-time straightening.} Rectified Flow \citep{liu2023flow} straightens
trajectories globally via reflow, a multi-stage procedure, and consistency-type objectives
distill straightness \citep{song2023consistency,yang2024consistency,salimans2022progressive}.
Concurrent work, Iso-FM \citep{isofm2026}, regularizes pathwise acceleration through a
self-guided finite-difference penalty, the same global objective family as our
\S\ref{sec:intervention}, with experiments at CIFAR-10 scale and no spatial analysis.
None of these works asks \emph{where} difficulty lives. That is this paper's subject.

\paragraph{Region-adaptive inference.} RAS \citep{liu2025region} allocates
diffusion-transformer compute from reactive output-change and attention signals. SDiT
\citep{sdit2026semantic} uses a ``spatial complexity'' heuristic, a reactive sampling-time
signal unrelated to our \signalname{} despite the name. Foveated and mixed-resolution
methods allocate by saliency \citep{upsample2025region,chao2026foveated}. None of these signals
is validated against per-region reference-solver discrepancy. \S\ref{sec:notlandscape} provides that
validation.

\paragraph{Curvature analyses.} Prior analyses relate global/average trajectory curvature
to sampling cost \citep{lee2023minimizing,luo2025curveflow,karras2022elucidating}. Our
unit of analysis is the region within a sample, not the whole trajectory. We validate
per-region against a near-exact reference solve, and the object we measure feeds back into a
training objective.

\section{\signalname{}: definition and grounding}
\label{sec:fc}

\paragraph{Setup.} A trained model induces a probability-flow ODE
\begin{equation}
\frac{dx}{dt} = f(x_t, t), \qquad x_{t_0}\sim\mathcal N(0,I)\ \text{(noise)} \;\to\; x_{t_1}\ \text{(data)}
\end{equation}
from noise to data, with PF-ODE drift $f$. \signalname{} is computed on a \emph{recorded field} $u$. For flow
matching $u = f = \vt$ is the learned velocity. For diffusion the drift $f$ \citep{song2021score} is an affine
function of the network's $\varepsilon$- or $x_0$-prediction, and we take that prediction as $u$, so
$u\neq f$ (the $\varepsilon$- and $x_0$-variants agree).
Because that affine map has time-varying coefficients, for diffusion we tie the variation of $u$ to
discretization error empirically rather than through the bound below.
We tokenize samples into regions
($16\times16$\,px regions at $256^2$ resolution). Writing $e = x^{\mathrm{cheap}}_T - x^{\mathrm{ref}}_T$ for
the endpoint error vector and $e_i = P_i e$ for its restriction to region $i$ ($P_i$ the region projection),
the \textbf{per-region integration error} is the within-region MSE $\lVert e_i\rVert^2$ between a cheap solve
(few-step Euler for flow, DDIM \citep{song2021ddim} for diffusion) and a near-exact reference solve from the
same noise. The reference is Heun-$200$ for the flow models and a fine DDIM ($50$--$100$ steps) for the
diffusion models (protocol in Appendix~\ref{app:details}). It is convergence-checked rather than assumed
exact: the per-region error map is essentially invariant under further refinement (rank correlation
$\ge 0.986$ between DDIM-$50$/$100$ and DDIM-$200$ on the diffusion models, and ${\approx}1$ for Heun-$200$
vs Heun-$400$ on the flow models), so it stands in for the converged endpoint. We call the target a
\emph{reference-solver discrepancy} throughout, not a ground-truth error. We use $e_i$ for this region-$i$ error vector throughout and
$\lVert e_i\rVert^2$ for its scalar magnitude.

\paragraph{Definition.}
\begin{definition}[\signalname{}]
\label{def:fc}
Over a trajectory window $W$ with drift sampled at spacing $\dt$,
\begin{equation}
\FC(i) \;=\; \sum_{t\in W}\big\lVert u_{t+\dt}(i) - u_t(i)\big\rVert
\end{equation}
where $u_t(i)$ is the field $u$ (velocity for flow, $\varepsilon$-/$x_0$-prediction for diffusion) restricted
to region (token) $i$ at trajectory point $x_t$.
\end{definition}
As the grid refines, $\FC(i) \to \int_W \lVert a(i,t)\rVert\,dt$, where
$a = du/dt = \partial_t u + (\nabla_x u)\,\dot x$ is the material derivative of $u$ along the trajectory
$\dot x$. For a flow model the trajectory obeys $\dot x = u$, so $a$ is the trajectory acceleration and this
makes $\FC$ the \textbf{per-region total variation of the learned velocity field}. We use rank statistics
throughout, which are invariant to the overall $\dt$ scale. Unless stated otherwise we sample $W$ along
the recorded reference trajectory, which separates the signal from coarse-grid noise but is not what a
deployed system has. \S\ref{sec:horizon} measures how much of the prediction survives when \signalshort{}
is computed from the cheap solve itself.

\paragraph{Numerical grounding.} Euler's local truncation error per step of size $H$ is
$\tfrac{H^2}{2}\lVert\ddot x\rVert + O(H^3)$, and $\ddot x = a$ is the same acceleration
$\FC$ accumulates. This is why we read $\FC$ as a difficulty measure. It is a motivation and not a
ranking proof, since global error runs through the state-transition operator and a magnitude bound
does not order regions.

\begin{proposition}[magnitude bound, informal]
\label{prop:bound}
To leading order the per-region endpoint error is bounded by a $\Phi$-weighted sum of every region's
accumulated acceleration, $\lVert e_i(T)\rVert \le \tfrac{H}{2}\sum_j c_{ij}\,D_j + O(H^2)$, with
$D_j = \lim \FC(j)$ and $c_{ij}$ the largest norm of the state-transition sub-block $\Phi_{ij}$ over the
window. When coupling is block-local this reduces to a bound driven by $D_i$ alone. In general $\FC_i$
bounds the truncation \emph{injected} at region $i$, which $\Phi$ then redistributes across regions.
Formal statement, assumptions, and proof in Appendix~\ref{app:proofs}.
\end{proposition}
\noindent A bound is not a ranking. Because $\Phi$ couples regions unless it is block-local, region
$i$'s \emph{final} error is not a function of $D_i$ alone, so whether the $\FC$-ordering matches the
error-ordering is an empirical question, not a corollary.

\paragraph{Why ranking is nontrivial.} $\FC$ sums magnitudes, but error accrues from the \emph{net}
change and is then carried through the state-transition operator. A magnitude-only statistic discards
three effects: directional cancellation (an oscillatory-but-returning region is over-scored), spatial
coupling, and nonuniform propagation across trajectory epochs. These perturb a ranking without
necessarily destroying it, consistent with correlations that are strong but below one, and the theory
does not derive $\rho$ values. Whether $\FC$ preserves the regional ranking is an empirical question,
which \S\ref{sec:landscape} answers against a near-exact reference solve.

\section{The difficulty landscape: four properties}
\label{sec:landscape}

All headline results are at ImageNet-256 scale on external state-of-the-art models
(SiT-XL/2, flow, latent \citep{ma2024sit}; DiT-XL/2, diffusion, latent
\citep{peebles2023scalable}; ADM, diffusion, pixel \citep{dhariwal2021diffusion}), with
our own smaller from-scratch models as controlled complements where retraining is
intrinsically required (Appendix~\ref{app:details}).

\subsection{Concentrated}
\label{sec:concentrated}
\begin{figure}[tbp]
\centering
\includegraphics[width=0.52\linewidth]{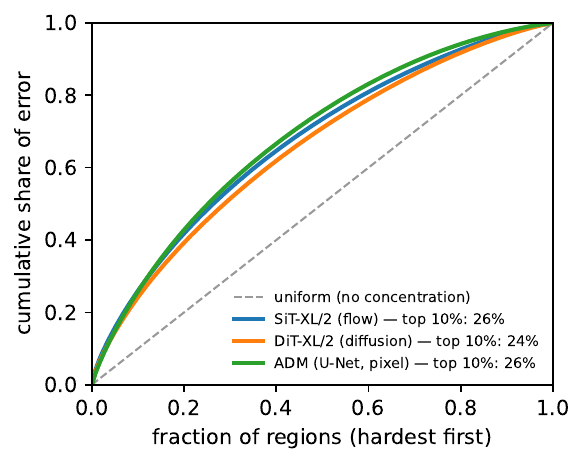}
\caption{\textbf{The landscape is concentrated.} Lorenz curves of per-region reference-solver
discrepancy (cumulative error share vs.\ fraction of regions, hardest first) for the three external
models; the dashed diagonal is uniform (no concentration). The top decile of regions carries
$24$--$26\%$ of the error across models, $2.4$--$2.6\times$ a uniform allocation, a moderate but
reproducible concentration.}
\label{fig:lorenz}
\end{figure}

Per-region integration error is moderately concentrated: the top 10\% of regions carry
\textbf{26.2\%} (SiT), \textbf{24.1\%} (DiT), \textbf{28.4\%} (ADM), \textbf{32.5\%}
(CelebA-HQ FM-DiT), and \textbf{26.3\%} (ImageNet FM-DiT) of total error (Gini
0.31--0.41). Lorenz curves in Fig.~\ref{fig:lorenz}. Both the moderate concentration and \signalshort{}'s
predictivity are stable across region granularity: on SiT at $8{\times}8$/$16{\times}16$/$32{\times}32$ grids
the top-decile share is $22$/$27$/$31\%$ and FC--error $\rho$ is $0.61$/$0.57$/$0.54$
(Appendix~\ref{app:multiscale}).

\subsection{Stable}
\label{sec:stable}
The landscape is a property of the model, stable against both the sampling budget and the training seed. Across cheap-solver
budgets (NFE 6 vs.\ 10), the error map correlates at $\rho = \textbf{0.88}$ (SiT) and
$\textbf{0.95}$ (ADM). Across full retraining with different seeds, testable only on
our controlled models, maps correlate at $\rho = \textbf{0.90}$--$\textbf{0.96}$, and
across model capacities (9.8M$\to$77M) at $\textbf{0.87}$--$\textbf{0.91}$.

\subsection{Learned}
\label{sec:learned}
\begin{figure}[tbp]
\centering
\includegraphics[width=\linewidth]{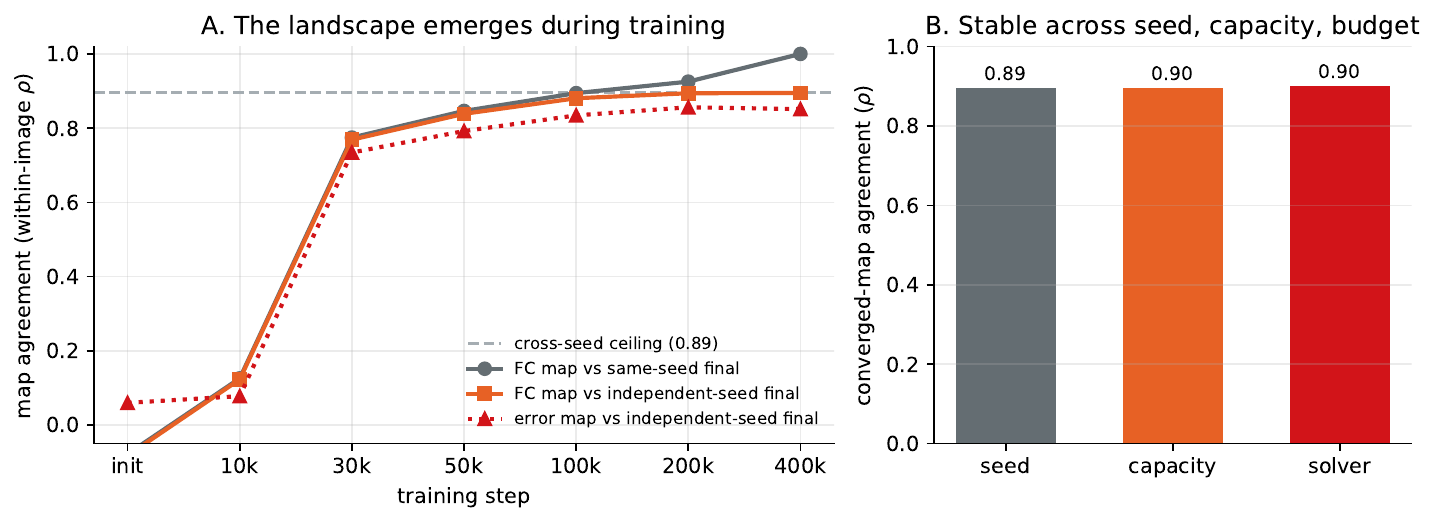}
\caption{\textbf{The difficulty landscape is learned and stable} (controlled CIFAR model; early-complexity
map; within-image Spearman on shared noise). \textbf{(A)} agreement between the map at training step $t$ and a
converged reference. Measured against an \emph{independently seeded} run's final map (orange, no
self-correlation), the map is uninformative at initialization and rises to the cross-seed ceiling
$\rho{\approx}0.90$ (dashed), so the identity of the difficult regions is learned during training rather than
present at init. The same-seed curve (grey) continues to $1.0$ only through self-correlation. The error map
itself (red, reference-solver discrepancy) emerges the same way, $\rho$ $0.06$ at initialization to $0.85$
cross-seed, so the target and not just the signal is learned structure. \textbf{(B)} the
converged map agrees at $\rho{\approx}0.90$ across independent seeds, model capacities, and solver budgets.}
\label{fig:stability}
\end{figure}

An untrained model's per-region \signalshort{} map does not correlate with the trained model's. Within-image
map-vs-final $\rho{\approx}0$ at initialization, rising to only ${\approx}0.13$ by $10$k steps. The
\signalshort{} map then sets in fast, reaching $\rho{\approx}0.84$ by $50$k (an eighth of the budget) and the
cross-seed ceiling ${\approx}0.90$ by $100$k, where it stays (controlled CIFAR model,
Fig.~\ref{fig:stability}). The reference-solver discrepancy map itself, not only the diagnostic that reads
it, shows the same emergence: uninformative at initialization (cross-seed $\rho{\approx}0.06$) and rising to a
cross-seed ceiling $\rho{\approx}0.85$ by $100$--$200$k. We call the landscape emergent, not absent at init,
because the initialized map is uninformative about the final one, not because we can prove it nonexistent. What training fixes is the
\emph{identity} of the difficult regions, and that identity is not set by the architecture or the
discretization alone. The frequency and edge controls of \S\ref{sec:notlandscape} point the same way. The
landscape belongs to the learned dynamics and is not reducible to the low-level image statistics we test.

\subsection{Predictable}
\label{sec:predictable}
\begin{figure}[tbp]
\centering
\includegraphics[width=0.5\linewidth]{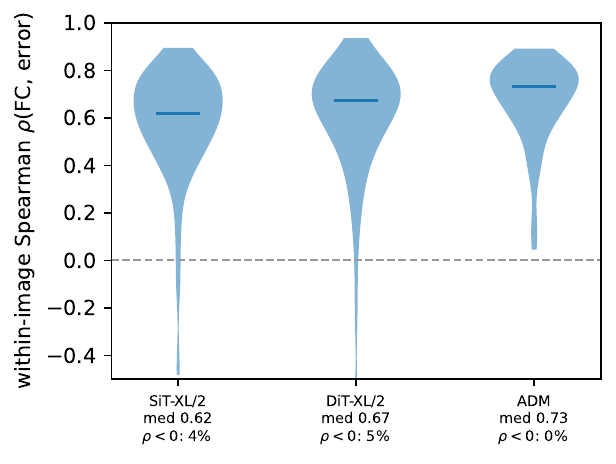}
\caption{\textbf{\signalshort{} predicts error for the \emph{typical} image, not just on average.}
Per-image within-image Spearman $\rho$ between \signalshort{} and the per-region reference-solver
discrepancy (violins; white bar = median) for the three external models. Medians ($0.62$--$0.73$) exceed
the means and only $0$--$5\%$ of images fall below $\rho = 0$ (dashed), so the aggregate is not carried by
a few high-correlation images.}
\label{fig:perimage}
\end{figure}

\signalshort{} computed from the model's velocities along the reference trajectory predicts the
per-region reference-solver discrepancy of a cheap solve: within-image $\rho = \textbf{0.586}$ (SiT),
$\textbf{0.574}$ (DiT), $\textbf{0.681}$ (ADM), $\textbf{0.70}$ (CelebA-HQ),
$\textbf{0.60}$ (ImageNet FM-DiT). This is the paper's central empirical claim, and every
number is validated against a near-exact solve, not a proxy. It holds \emph{per
image}, not merely on average (Fig.~\ref{fig:perimage}). The within-image $\rho$ has a median of
$0.62$--$0.73$ across the external models, \emph{above} its mean, with only $0$--$5\%$ of images
falling below $\rho = 0$, so \signalshort{} is reliable for the \emph{typical} image and not carried
by a few. Image-level bootstrap $95\%$ CIs (resampling images, the correct inference unit given spatial
autocorrelation) are $[0.54, 0.62]$ (SiT) and $[0.63, 0.71]$ (ADM).

\subsection{What the landscape is not}
\label{sec:notlandscape}
Per-region image frequency and edge magnitude, the intuitive explanations and the
signals region-adaptive methods actually deploy, do not explain it. In latent space,
frequency and edges are uncorrelated with error ($\rho = -0.03$/$-0.13$, SiT) and
partialling them out leaves \signalshort{}'s prediction intact (partial
$\rho = 0.747$/$0.761$). In pixel space they correlate (${\approx}0.51$, ADM), but
\signalshort{} beats and survives them ($0.765 \to 0.662$/$0.679$ partial). In a
matched-budget selector comparison (controlled study; Appendix~\ref{app:details}),
\signalshort{} is the only signal that outperforms random selection. Edge, frequency,
attention-entropy, and velocity-magnitude selectors do not. The FID-optimal selection is the reverse of
the error-optimal one, so \signalshort{} marks where a solve deviates numerically, not where correcting it
most improves sample quality (Appendix~\ref{app:selector}).
We further reproduce the actual signals region-adaptive samplers use: the SDiT complexity of
\citet{sdit2026semantic} (edge/Laplacian/residual on the predicted clean image, Eq.~5) and
RAS's per-step output change. At a \emph{matched} early observation window both score
$\rho \lesssim 0.15$ against \signalshort{}'s $0.60$ ($n=512$, SiT); the SDiT signal reaches
\signalshort{}'s level ($0.58$) only through its residual term read at ${\sim}90\%$ of
sampling, reactively, once most of the trajectory is already computed. \signalshort{}'s
advantage is therefore \emph{early} predictivity, which these reactive signals do not provide. This scores
the signals as \emph{early predictors} of the final discrepancy, the use \signalshort{} is built for; it is
not a comparison of the RAS or SDiT samplers at their own reactive allocation task.

Against numerical rather than perceptual alternatives, \signalshort{} is representative of its family, not
uniquely optimal, and we report this plainly. On a uniform fine grid the accumulated embedded
Euler--Heun local-error estimate is, to leading order, $\tfrac{h}{2}\sum\lVert\Delta u\rVert$,
\signalshort{} up to a constant, so the solver-native estimator is equivalent rather than a separate
baseline at either grid (the coarse-grid version is the cheap-trajectory signal of \S\ref{sec:horizon}). Distinct variation statistics from the
same record at the same windows (signed cumulative change $\lVert\sum\Delta u\rVert$, squared variation,
max step, horizon-weighted variation; $n{=}96$ per model, fresh seeds) land close to or somewhat above
\signalshort{}. At the early window the signed variant reaches $\rho = 0.73$ against \signalshort{}'s
$0.65$ on SiT and $0.75$ against $0.73$ on CelebA, and a horizon-weighted variant is the best full-window
predictor on both. This is consistent with the transport account, since signs and remaining horizon govern
how injected defects reach the endpoint (\S\ref{sec:transport}), while velocity magnitude stays far behind
($0.19$--$0.25$ early). We keep \signalshort{} as the reported diagnostic because it is the member the
Euler bound grounds and the one the intervention of \S\ref{sec:intervention} penalizes. The family-level
conclusion, that trajectory variation read early predicts where a cheap solve errs, does not depend on
the choice (Appendix~\ref{app:numbase}).

\section{When is difficulty readable? The prediction horizon}
\label{sec:horizon}

\begin{figure}[tbp]
\centering
\includegraphics[width=0.62\linewidth]{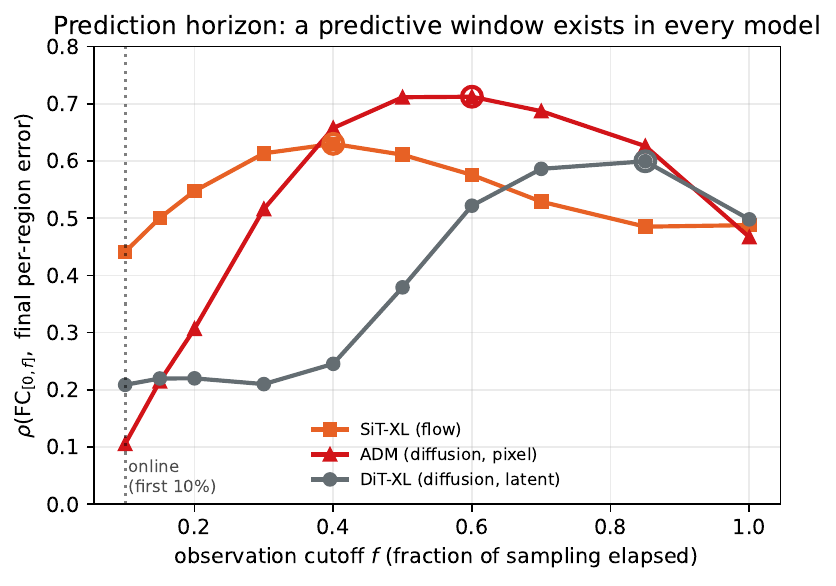}
\caption{\textbf{Prediction horizon.} Within-image correlation between early-window \signalshort{}
(observed up to cutoff $f$) and the final per-region integration error, for SiT (flow), ADM
(diffusion, pixel), and DiT (diffusion, latent). Every model exposes a predictive window (open
circles mark each peak); its location depends on the model, early for the flow model (already
$\rho=0.44$ at $f=0.1$) and later for the diffusion models. \emph{Where} error is predictable varies.
\emph{That} it is predictable is what holds across every model.}
\label{fig:horizon}
\end{figure}

Sweeping the observation cutoff $f$ (fraction of sampling elapsed) yields each model's
readability profile (Fig.~\ref{fig:horizon}). A predictive window exists in every model,
and its location depends on the model, early for the flow model (SiT: $\rho = 0.44$ at
$f = 0.1$, peaking $0.63$ at $f = 0.4$) and later for the diffusion models (ADM peaks $0.71$
at $f = 0.6$, DiT $0.60$ at $f = 0.85$). The window coincides with the measured
\emph{error-determining window} $W^*$, the trajectory epoch whose acceleration
correlates most strongly with the final per-region error (per-bin
acceleration$\to$final-error correlation: SiT strongest earliest, $0.55$; ADM mid,
$0.60$; DiT mid/late, $0.51$; Appendix~\ref{app:proofs}). The horizon and the
error-determining epoch are thus the same measurement. \emph{Where} error is predictable varies.
\emph{That} it is predictable is what holds across every model. To be plain about what this
buys: on DiT the window opens only after most of sampling has elapsed, which supports
late-stage error accounting rather than early warning. The early-localization payoff below
is a flow-model result.

\paragraph{Payoff: early error localization.} On SiT, flagging the
top-15\% of regions by \signalshort{} accumulated over the first 30\% of sampling captures
\textbf{27.9\%} of the final error, \textbf{82\%} of the oracle ceiling (34.2\%) and
$1.9\times$ the random baseline. The signal is causal in time, seeing neither the rest of
the trajectory nor the final error, but as reported it is accumulated on the fine reference
grid. A deployed system has only the coarse trajectory it is integrating, so we also
recompute \signalshort{} from the NFE-6 Euler solve itself on held-out seeds. On an
exploratory seed, within-image $\rho$ reaches $+0.32$ from the first two velocity
evaluations and $+0.26$ from all six, against $+0.58$ for the early fine-grid signal on
the same images. Because that prefix was chosen after seeing the results, we froze it and
the metrics and ran a confirmatory seed: $\rho = +0.40$ $[0.36, 0.44]$, and flagging the
top-15\% of regions by the two-evaluation cheap prefix captures $24.5\%$ $[23, 26]$ of the
final error, versus an exact random share of $14.8\%$ and an oracle ceiling of $35.2\%$.
This is $70\%$ of the raw oracle capture, or $47\%$ of the improvement from random to oracle
(Appendix~\ref{app:details}). Under the same frozen protocol the result replicates on our
own CelebA-HQ-256 model (capture $26.2\%$ $[25, 28]$, $\rho = +0.46$). The cheap-trajectory
signal is weaker than the fine-grid one, but it localizes error from the first third of a
six-step solve, on both models tested.

\section{Spatial transport of integration error}
\label{sec:transport}

Difficulty has so far been local: a per-region signal for a per-region error. Proposition~\ref{prop:bound}
says that cannot be the whole story. The final per-region error
$e_i(T)\approx\sum_k\sum_j\Phi_{ij}(T,t_k)\,\tau_k^{(j)}$ (to first order; nonlinear remainders enter at
higher order) sums the truncation $\tau_k^{(j)}$ injected at each
region $j$ (its magnitude tracked by $\FC_j$) after the state-transition operator $\Phi$ has moved it
around. The theory leaves $\Phi$ abstract. We do not recover the operator itself; we probe the directional source-to-destination influence it induces (through the endpoint Jacobian), ask how far injected error travels, and then
rebuild $e_i$ from its parts to see how much of a region's final error is its own and how much arrives from
elsewhere.

\paragraph{Perturbation protocol.} At trajectory time $t^\ast$ we perturb source region $j$ by a fixed-norm
vector $z_j$ supported on region $j$ ($\lVert z_j\rVert=\varepsilon$), integrate both $\pm$ perturbed states
to the data endpoint with the same near-exact solver, and read the endpoint response per region by central
difference,
\[
M_{ij}(t^\ast)=\frac{\lVert P_i\,(x_T^{\,j,+}-x_T^{\,j,-})\rVert}{2\varepsilon}
\;\approx\; \frac{\lVert \Phi_{ij}(T,t^\ast)\,z_j\rVert}{\lVert z_j\rVert},
\]
a norm-valued regional transport kernel. Everything is measured in the model's own state space, latent for
the flow transformers and pixel for the U-Net, so dynamical coupling is never confounded with a decoder's
receptive field. We summarize $M$ by the self-retained response $R_{\mathrm{self}}=M_{jj}/\sum_i M_{ij}$, the
share of the summed regional response magnitude that stays in the source region. It is a dispersion
index, not a conserved mass. We report image-level medians with bootstrap $95\%$ CIs
(SiT $n{=}32$, CelebA $n{=}48$, ADM $n{=}16$). An orthogonally additive energy split gives the same
picture with more weight on the source: near noise the source region retains $0.10$--$0.40$ of the
response energy, while the whole response spans $9$--$43$ effective regions
(Appendix~\ref{app:energysplit}).

\begin{figure}[tbp]
\centering
\includegraphics[width=\linewidth]{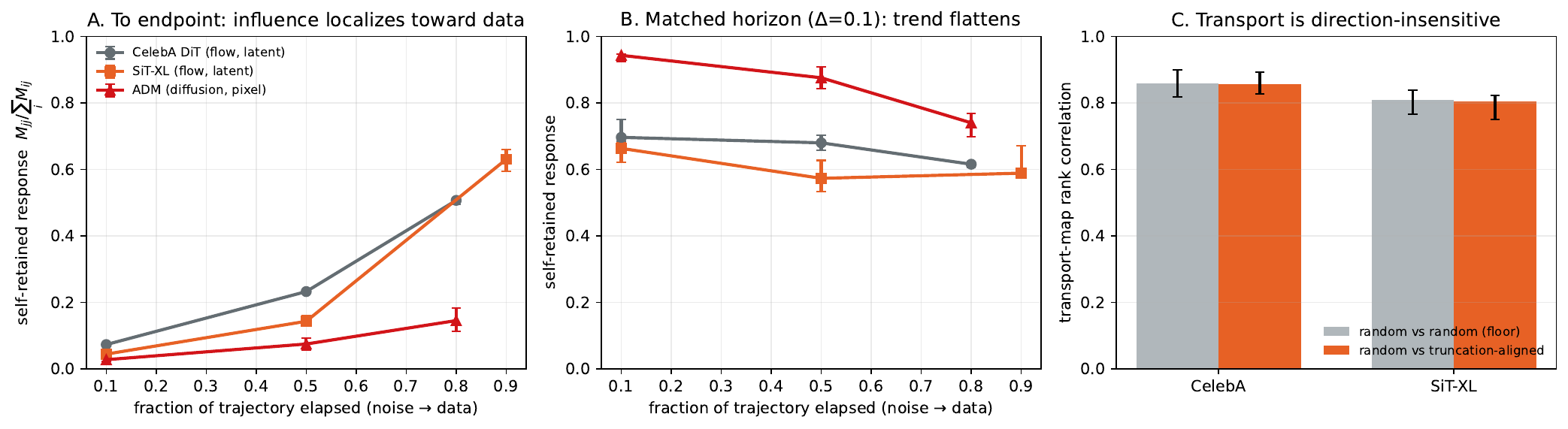}
\caption{\textbf{Integration error is spatially transported; its apparent temporal localization is a horizon
effect.} A region is perturbed at trajectory time $t^\ast$ and the endpoint response is read per region.
\textbf{(A)} to the endpoint, the self-retained response rises toward data on a flow transformer (SiT-XL), a
diffusion U-Net (ADM, pixel), and our own model. \textbf{(B)} under a matched rollout horizon
($\Delta=0.1$) the trend flattens or reverses, so the localization in (A) is horizon-driven. \textbf{(C)} on
the flow models, the truncation-aligned transport map matches the random-direction map at the reproducibility
floor. Bands in (A, B) are bootstrap 95\% CIs of the image-level median; (C) shows the $2.5$--$97.5$th
percentile interval over pairs. The endpoint profiles in (A) use SiT $n{=}32$, CelebA $n{=}48$, ADM
$n{=}16$; the horizon control (B) and direction control (C) use $n{=}8$. The map survives solver
refinement (rank correlation $\ge 0.97$).}
\label{fig:transport}
\end{figure}

\paragraph{A local disturbance spreads.} On every model, a perturbation confined to one region influences
distant endpoint regions. Near the noise end almost none of the response stays home. The self-retained
fraction is $0.05$ (SiT), $0.07$ (CelebA), $0.03$ (ADM), at a radius of $0.74$--$0.85$ of the
fully-delocalized ceiling (Fig.~\ref{fig:transport}A). ADM, a pixel-space convolutional U-Net with no VAE
and an architecture entirely unlike the latent transformers, delocalizes as strongly, so the effect is not
an artifact of latent patchification or of the transformer backbone. On our own model the transport is absent at initialization (self-retained $1.0$, a near-identity
flow) and emerges during training, strengthening from $0.31$ at $10$k steps to $0.23$ at $100$k, so the
architecture alone does not produce it. Refining the reference solver leaves the map essentially unchanged (rank correlation $1.00$ on
SiT, $0.97$ on ADM), the response is linear in $\varepsilon$ across a factor of four in perturbation scale,
and it holds across solver families.

\paragraph{Endpoint influence contracts toward data, but only because the horizon shrinks.} As $t^\ast$
approaches the data end, $R_{\mathrm{self}}$ rises steeply (to $0.63$ SiT, $0.51$ CelebA, $0.15$ ADM), so
endpoint influence looks progressively more local. It is tempting to read this as the dynamics becoming
local near data. A matched-horizon control shows they do not. Rolling out a fixed trajectory fraction
($\Delta=0.1$) from every $t^\ast$, rather than all the way to $T$, flattens or reverses the trend on all
three models ($R_{\mathrm{self}}$ from noise to data: CelebA $0.70$ to $0.62$, SiT $0.66$ to $0.59$, ADM
$0.94$ to $0.74$; Fig.~\ref{fig:transport}B). The localization in (A) is the shrinking remaining horizon,
not an intrinsic tightening of the per-step coupling, which if anything strengthens toward data.
Endpoint influence contracts along the trajectory, but the matched short-horizon influence does not.

\paragraph{The perturbation direction barely matters (flow models).} Because $M$ uses a chosen direction, we
check it against the geometry of true truncation error. Perturbing along the local acceleration, the
leading-order truncation direction, gives a transport map that agrees with the random-direction map at the
reproducibility floor (rank correlation $0.81$, essentially the random-versus-random value; SiT, $n{=}26$).
Generic perturbations therefore produce regional influence rankings similar to those of actual truncation
residuals, on the tested flow models. On the
diffusion U-Net the same control is noisier and its confidence interval overlaps the floor, so we make this
claim only for the flow models.

\begin{figure}[tbp]
\centering
\includegraphics[width=\linewidth]{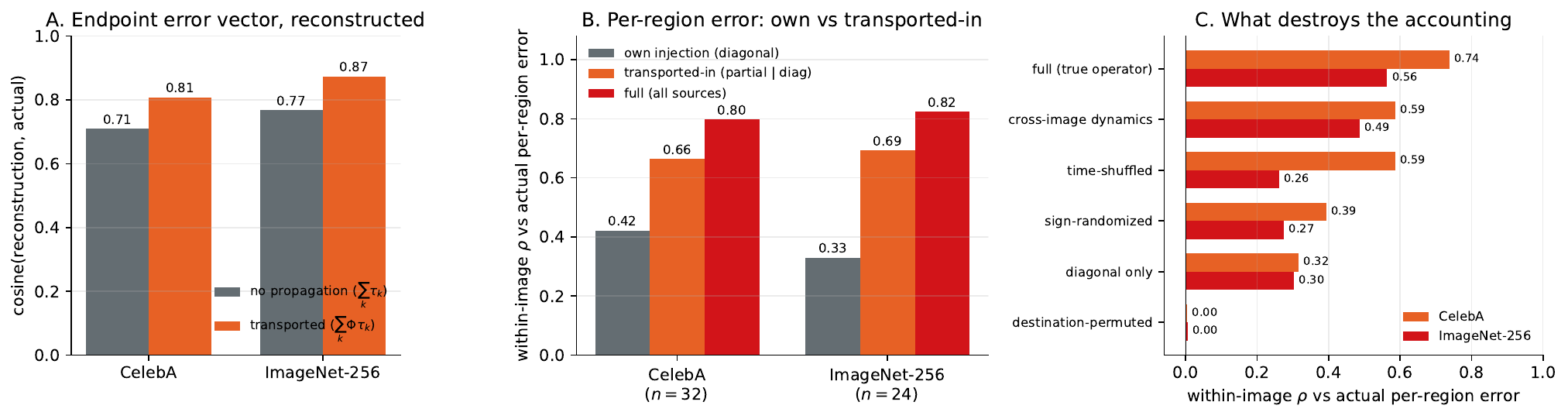}
\caption{\textbf{Reconstructing endpoint error from signed truncation residuals} (our own $256$px models;
post-hoc, using the true residuals and the model's Jacobian). \textbf{(A)} the endpoint error \emph{vector},
rebuilt by summing the residuals with vs.\ without propagation through the linearized dynamics: propagation
lifts the cosine from $0.71/0.77$ to $0.81/0.87$. \textbf{(B)} per-region error, within-image $\rho$ against
the actual per-region error: a region's own injected error (diagonal) explains only part ($\rho\approx0.33$--$0.42$),
the error transported in from other regions is more strongly associated with it (partial
$\rho\approx0.66$--$0.69$ given the diagonal), and the full source-summed reconstruction tracks it closely
($\rho\approx0.80$/$0.82$). \textbf{(C)} the operator-null ladder, protocol frozen before evaluation
($n{=}16$ per model; $15$ cross-image pairs):
scrambling \emph{where} each source's contributions land, with content, norms, and signs intact,
annihilates the reconstruction on both models. Sign randomization more than halves it. Shuffled step
timing and another image's dynamics cost less, so the operator's spatial structure is partly shared
across images. Correspondence carries the accounting.}
\label{fig:reconstruction}
\end{figure}

\paragraph{Rebuilding a region's error from its sources.} A decomposition is only worth the name if it holds
quantitatively. This is a post-hoc mechanistic account, not a deployable predictor like \signalshort{}. It
uses the actual truncation residuals and the model's own Jacobian, so the question is not whether it predicts
error but whether the first-order source-and-transport account closes. On our own models, where every source
is accessible, we take the actual signed per-step truncation $\tau_k$, the Euler step minus the accurate step
evaluated at the reference state, propagate each region's share of it to the endpoint through the model's own
linearized dynamics, and add the contributions back up. The reconstruction matches the direction of the true endpoint
error at cosine $0.81$ on our unconditional CelebA model and $0.87$ on a class-conditional ImageNet-256 model,
and recovers its overall magnitude too (norm ratio $0.98$ and $1.01$; Fig.~\ref{fig:reconstruction}A).
Without propagating the residuals these fall to $0.71$ and $0.77$. Signed residual accumulation already
captures much of the endpoint direction, and propagation through the learned dynamics adds a substantial,
consistent increment on both. The increment is not a coarse-grid artifact. As the cheap step shrinks,
the unpropagated sum degrades (cosine $0.81$ to $0.53$ across NFE $4$ to $16$) while the transported
reconstruction holds, so propagation matters more exactly where first-order theory says it should
(Appendix~\ref{app:recon}). Split by region, a region's own injected error, propagated to the
endpoint and read back where it started, explains only part of that region's final error (within-image
$\rho \approx 0.33$ on ImageNet, $0.42$ on CelebA). What arrives from other regions is more strongly
associated with a region's final error, at within-image partial rank correlation $\approx 0.66$--$0.69$
(median over images; CelebA $n{=}32$, ImageNet $n{=}24$) once its own injection is controlled
for. Regional rank statistics are evaluated on $64$ strided destinations
out of the $256$ regions. The full source-summed reconstruction reaches within-image $\rho \approx 0.80$
and $0.82$ against the
actual per-region error (Fig.~\ref{fig:reconstruction}B). This depends on more than the magnitudes of the incoming
contributions: preserving every contribution's norm while randomizing their relative signs substantially
reduces the reconstruction (within-image $\rho\,0.27$--$0.39$, no better than a region's own injection,
versus $\rho\,0.56$--$0.74$ for the correctly-signed sum; Appendix~\ref{app:recon}). The correct signed
superposition of the propagated residuals therefore carries information beyond how much energy reaches each
region. A family of operator nulls, protocol frozen before evaluation and run on the $64$-source
protocol below (whose full reconstruction is $\rho\,0.74$/$0.56$; the all-source full-field number
above is $0.80$/$0.82$), sharpens what carries the closure
(Appendix~\ref{app:recon}). Reassigning each source's contributions among destinations, keeping every
block's content, norms, and signs, destroys the reconstruction outright ($\rho = 0.00$ on both models,
below even the sign null). Injecting the residuals at shuffled step times degrades it ($0.74$ to $0.59$
on CelebA, $0.56$ to $0.26$ on ImageNet), while propagating them through another image's dynamics costs
less ($0.59$ and $0.49$). Where a contribution lands is the most load-bearing ingredient of the
accounting, and the operator's spatial structure is partly shared across images of the same model, which
matches the landscape being learned model-level structure (\S\ref{sec:learned}). We claim correspondence
and sign specificity, not image specificity. This is the concrete form of ``injected locally,
transported nonlocally,'' and it is why $\FC$, a purely local magnitude, tracks final error only in part.

\paragraph{Scope.} The reconstruction and decomposition run on our own two models. On the large external
models we establish the transport measurement, cross-architecture and solver-verified, but not the
reconstruction. Transport is a statement about the dynamics. Whether a local training penalty is
therefore suboptimal turns on optimization-space coupling we do not measure.

\section{Intervening on \signalname{}}
\label{sec:intervention}

\paragraph{Objective.} We add a penalty on the model's own local velocity variation
(\methodname{}, \methodshort{}):
\begin{equation}
\label{eq:fcm}
\mathcal L \;=\; \mathcal L_{\mathrm{FM}} \;+\; \lambda\,
\big\lVert \vt\big(x_t + \dt_{\mathrm{eff}}\,\bar v,\; t+\dt\big) - \vt(x_t, t)\big\rVert^2,
\end{equation}
where $\bar v$ is the detached current velocity (a self-guided Euler lookahead),
$\dt_{\mathrm{eff}}$ handles the time boundary per sample, and for conditional
training both evaluations share one classifier-free-guidance drop mask, so the penalty
measures trajectory change and not conditioning change. Cost: one additional forward
evaluation per step (${\approx}2\times$ step time at equal batch; no additional
parameters, no teacher, no reflow stage). $\lambda = 0.3$ throughout (selected by a
controlled sweep; larger $\lambda$ over-straightens and taxes the high-NFE ceiling).

\paragraph{Framing, and relation to concurrent work.} We use a finite-difference
\signalshort{} regularizer as our intervention, a controlled test of whether
reducing the diagnosed quantity changes few-step integrability and sample quality. Concurrent Iso-FM
\citep{isofm2026} independently studies a related global acceleration penalty. Our
contribution is not the penalty in isolation but the spatial characterization and its
mechanistic validation. Here we establish the intervention's effect on the
\emph{spatial structure} identified in \S\ref{sec:landscape}, at 256px and ImageNet
scale. \methodshort{} reduces measured \signalshort{}, the basic mechanism check. It cuts
reference-solver discrepancy by 57--65\% at fixed budget. The relative error reduction rises
with baseline difficulty percentile, from $-58\%$ in the easy half to $-72\%$ at the 90--95th
percentile (Fig.~\ref{fig:mechanism}), so it is largest in the regions \S\ref{sec:landscape} flags
as difficult, though regression to the mean contributes. And the top-decile error share falls
($26.9\% \to 22.9\%$), so the tail is compressed, not merely shifted.

\section{Results: from-scratch training at 256px}
\label{sec:results}

\methodshort{} here is a construct-validity intervention, not a proposed sampler or a compute-efficient
training method. It asks whether the numerical structure we measured can be moved by training: it lowers
\signalshort{} and reference-solver discrepancy and improves few-step FID at matched optimization steps, but
it does \emph{not} win at matched training compute on ImageNet (below). Read the FID numbers in that light;
the full tables, $\lambda$-sweep, seed variance, and compute ladders are in Appendix~\ref{app:details}.

\paragraph{Protocol.} Paired arms (baseline vs.\ \methodshort{}) trained from scratch in lockstep:
identical architecture, data order, noise, timesteps, and conditioning masks, differing only by the
\methodshort{} gradient, at matched optimization steps and data exposure. This is not matched training
compute: \methodshort{} adds one forward per step (measured $1.9\times$ step time), which is why the
compute-matched comparisons below carry the verdict. Compact single-GPU models (33--77M) at these
schedules do not target leaderboard FID; we report paired deltas and state absolute numbers plainly
(Appendix~\ref{app:details}).

\begin{figure}[tbp]
\centering
\includegraphics[width=0.92\linewidth]{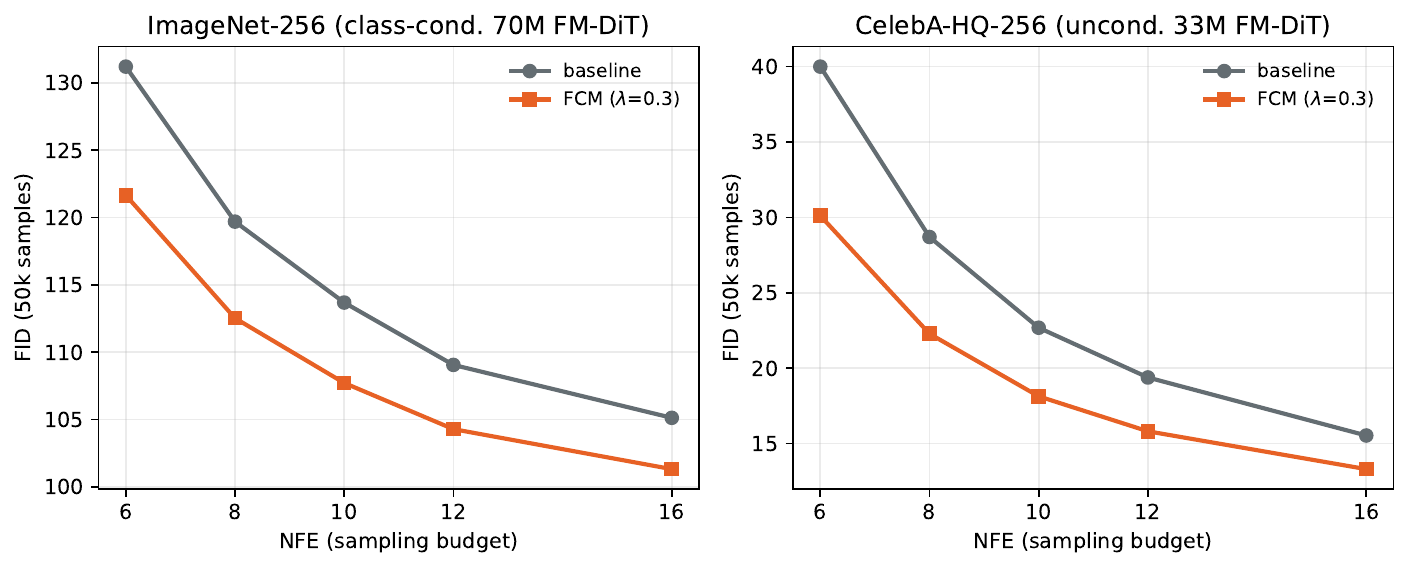}
\caption{\textbf{Minimizing \signalname{} improves few-step FID at matched optimization steps (256px).}
FID (50k samples, cfg $=1.0$) vs.\ sampling budget (NFE), baseline vs.\ \methodshort{} ($\lambda=0.3$),
trained from scratch in lockstep on class-conditional ImageNet-256 (left) and CelebA-HQ-256 (right).
The gain is largest in the few-step regime and present at every NFE we test, with no high-NFE cost.
The compute-matched verdict differs by scale (see text).}
\label{fig:fid}
\end{figure}

\paragraph{Matched optimization steps.} On class-conditional ImageNet-256 (70M FM-DiT, 100k steps),
\signalshort{} falls $5.3\%$, integration error $57\%$, and FID improves at every budget, most in the
few-step regime ($131.2 \to 121.7$ at NFE 6, $-7.3\%$; Table~\ref{tab:imagenet},
Appendix~\ref{app:e4}), persisting under stronger guidance (cfg $=1.5$: $-8.7\%$ to $-3.2\%$). On
unconditional CelebA-HQ-256 (33M), integration error falls $65\%$ and FID improves at every budget
($40.0\to30.1$ at NFE 6), with no diversity loss on same-noise grids (Fig.~\ref{fig:fid}). A convergence
test specified in advance trains both arms to $4\times$ the schedule; the step-matched NFE-6 delta grows
from $4.4\%$ to $11.4\%$ rather than vanishing, so the gain is not an undertraining artifact
(Appendix~\ref{app:e4}).

\paragraph{Matched training compute, the stricter test.} On CelebA we match forward-equivalent compute
(\methodshort{} at $S$ steps against the baseline at $2S$, which favors the baseline) and \methodshort{}
still wins the few-step regime ($34.7$ vs.\ $41.3$ at NFE 6, larger gaps on the integration error it
targets directly). At ImageNet scale the advantage does not survive the correction: the baseline at
$200$k steps, already more compute than \methodshort{} at $100$k, reaches NFE-6 FID $118.7$ against
$124.6$, with a larger margin at NFE $10$ and $16$. Following the advance-specified rule we report
\methodshort{} on ImageNet as a mechanism with better progress per update, though not per unit of
training compute, and we do not claim the compute-matched few-step win as a general property. It holds
at CelebA scale and not at ImageNet scale (full ladders in Appendix~\ref{app:e4}).

\section{Limitations and scope}
\label{sec:limitations}

\methodshort{} is studied as a \textbf{from-scratch training objective}; other training
regimes (e.g., adapting already-converged models) are outside our claims. \textbf{Training
variance.} The 256px intervention arms are a single training run per cell; the 50k-sample
FID and the image-level bootstrap intervals we report bound \emph{evaluation} noise, not
\emph{training-run} variance. On the controlled $32\times32$ model, where replication is
affordable, we train three independently seeded paired runs (baseline vs.\ \methodshort{},
matched seed and data order, $50$k steps each). The few-step improvement holds across seeds. The
paired $\Delta$FID at NFE $6$ is $+6.97 \pm 3.26$ with all three seeds favoring
\methodshort{}, and $+2.87 \pm 2.34$ at NFE $8$ (again all three). By NFE $10$--$16$ the
paired delta falls within seed noise, consistent with the controlled-model gain being a
few-step effect (Appendix~\ref{app:seedvar}; matching the $\lambda$-sweep, where the
$32\times32$ gain also concentrates at low NFE). The 256px intervention results are therefore paired controlled evidence, not a variance-resolved
performance comparison. The controlled-model replication and consistency across three datasets
make a single-run fluke unlikely. \textbf{Compute.} \methodshort{}
adds one forward per step (measured $1.9\times$ step time); our deltas are at matched
optimization steps and data exposure, not matched training compute, and the E4 study
tests persistence as the baseline is trained to $4\times$ the steps. Our from-scratch
models are compact and short-scheduled relative to leaderboard systems: paired deltas are
the claim, not absolute state-of-the-art quality. \textbf{What the metric is.} Our target
is per-region \emph{solver discrepancy} (cheap vs.\ near-exact integration), not
perceptual importance. Correcting high-\signalshort{} regions reduces pixel or latent
error, which need not coincide with perceptual salience. The \signalshort{}$\to$error
coupling is monotone but not proportional, so we claim integrability improvement, not
proportionality. The prediction horizon's location varies by model, and we claim its
existence and a protocol to measure it, not universality of ``early''. The
``error reduction concentrates on hard regions'' claim is reported per fixed
\signalshort{}-decile and is subject to regression-to-the-mean, so we treat it as descriptive
of where gains appear, not as an independent effect. The $32\times32$ ablations predate a
minor penalty-implementation refinement (shared conditioning masks) that all 256px runs
include.

\section{Conclusion}
\label{sec:conclusion}

Numerical difficulty in learned generative flows is an identifiable spatial property:
concentrated, stable, learned, and predictable from the model's own velocity or prediction field via
\signalname{}, a diagnostic of the local truncation the Euler residual injects. But
difficulty does not stay where it is injected. A perturbation analysis shows the learned dynamics
spread a local state disturbance across regions, and a signed-residual reconstruction shows actual
discretization defects take the same pathway, so final regional error is local injection
redistributed by propagation, which is why a local signal predicts it only in part.
Reading the landscape localizes error early, before a sample is finished. Computed from the
coarse solve alone the signal weakens ($\rho$ $0.62$ to $0.40$) yet protocol-frozen
confirmatory tests on two models still find it flagging regions that hold about a quarter
of the final error against an exact random share of $14.8\%$. A supporting
intervention that reduces \signalshort{} also reduces reference-solver discrepancy, with its
compute-adjusted effect on sample quality evaluated separately (\S\ref{sec:results}). We
expect this account of where error is injected and how it spreads to inform future decisions,
from per-sample step budgets to distillation targets, though we have not shown the coarse-solve
signal is strong enough to drive them. The online signal may support numerical confidence estimation
or adaptive solver diagnostics, but its relationship to perceptual quality remains unresolved.

\bibliographystyle{plainnat}
\bibliography{refs}

\appendix
\section{Proof and reconstruction details}
\label{app:proofs}

\subsection{Formal statement and proof of Proposition~\ref{prop:bound} (magnitude bound)}
\textbf{Formal statement.} Let $D_i := \int_W \lVert a_i(t)\rVert\,dt = \lim \FC(i)$ be region $i$'s
accumulated acceleration, and let $\Phi(T,t)$ be the state-transition operator of $J=\nabla_x u$ with
region sub-blocks $\Phi_{ij}$ (the variational linearization along the trajectory; nonlinear remainders
enter the $O(H^2)$ terms). With $u\in C^2$ on $W$ with bounded first and second derivatives
($L$-Lipschitz, so trajectories are $C^3$ in $t$ and the Euler remainder is genuinely $O(H^3)$ per
step), explicit Euler steps of size $H$, and $W=[t_0,T]$ the full integration interval,
\begin{equation*}
\lVert e_i(T)\rVert \;\le\; \tfrac{H}{2}\sum_j c_{ij}\,D_j \;+\; O(H^2),
\qquad
\lVert e(T)\rVert \;\le\; e^{L|W|}\tfrac{H}{2}\sum_j D_j \;+\; O(H^2),
\end{equation*}
where $c_{ij} := \sup_{t\in W}\lVert \Phi_{ij}(T,t)\rVert \le e^{L|W|}$; block-local coupling
($c_{ij}\approx0$ for $j\neq i$) reduces the first bound to
$\lVert e_i(T)\rVert \le e^{L|W|}\tfrac{H}{2}D_i + O(H^2)$. The directional-cancellation failure mode of
\S\ref{sec:fc} is the gap in
$\lVert\sum_{t\in W}\Delta u_i\rVert \le \sum_{t\in W}\lVert\Delta u_i\rVert = \FC(i)$, with equality
only for co-directional changes.
Euler recursion with step $H$: $e_{k+1} = (I + H J_k)e_k + \tau_k$ with $J=\nabla_x u$ and
local truncation $\lVert\tau_k\rVert \le \tfrac{H^2}{2}\lVert a(t_k)\rVert + O(H^3)$.
Unrolling to leading order, $e(T) \approx \sum_k \Phi(T,t_k)\,\tau_k$ (the variational linearization;
nonlinear remainders enter at higher order) where $\Phi$ is the state-transition
operator of $J$; Gr\"onwall gives $\lVert\Phi\rVert \le e^{L|W|}$ for $L$-Lipschitz $u$, and
$\sum_k H\lVert a(t_k)\rVert \to \int_W\lVert a\rVert\,dt$. Summing over $k$ gives the
\emph{full-state} bound $\lVert e(T)\rVert \le e^{L|W|}\tfrac{H}{2}\sum_j D_j + O(H^2)$.
For a single region, split each step's truncation by region, $\tau_k=\sum_j\tau_k^{(j)}$
with $\lVert\tau_k^{(j)}\rVert\le\tfrac{H^2}{2}\lVert a_j(t_k)\rVert$; then
$e_i(T)=\sum_k\sum_j\Phi_{ij}(T,t_k)\,\tau_k^{(j)}$, and taking norms with
$c_{ij}:=\sup_t\lVert\Phi_{ij}(T,t)\rVert$ gives
$\lVert e_i(T)\rVert\le\tfrac{H}{2}\sum_j c_{ij}D_j+O(H^2)$. The clean per-region form
$\lVert e_i(T)\rVert\le e^{L|W|}\tfrac{H}{2}D_i$ holds only under block-local coupling
($c_{ij}\!\approx\!0$, $j\neq i$). In general $\FC_i$ controls region $i$'s
\emph{locally injected} truncation, which $\Phi$ redistributes across regions. This
redistribution, together with the directional cancellation below, is exactly why the
per-region \emph{final}-error ranking is established empirically (\S\ref{sec:predictable}),
not derived here. \qed

\paragraph{What this does not give.} A bound is not a ranking: a region $A$ with bound
100/actual error 2 vs.\ a region $B$ with bound 50/actual error 40 has $\FC$ rank $A>B$
but true rank $B>A$. ``$\FC$ predicts the per-region error ranking'' is therefore an
\emph{empirical} result (\S\ref{sec:predictable}), motivated by Proposition~\ref{prop:bound}
but not derived from it. Likewise the theory cannot derive $\rho$ values; the
decomposition below bounds how much ranking a residual of a given size can scramble, but
it neither derives the residual's size nor the measured correlations.

\subsection{Mechanistic reconstruction: definitions and algorithm}
\label{app:recon}
Proposition~\ref{prop:bound} gives, to leading order, $e(T)=\sum_k\Phi(T,t_{k+1})\,\tau_k$ (dropping higher-order defect interactions). We test this first-order reconstruction directly on
our own models, where the full state is accessible. This is a post-hoc account, not a deployable predictor:
it uses the actual residuals and the model's own dynamics.

\paragraph{Reference trajectory and residuals.} Fix noise $x_1$. The reference states $x^{\mathrm{ref}}_k$ at
the cheap gridpoints $t_k$ ($k=0,\dots,K$) are read from a near-exact Heun solve on the fine grid
($K\!\cdot\!M$ steps, $K\!\cdot\!M\!\approx\!200$). The signed per-step truncation is the teacher-forced
one-step defect evaluated \emph{at the reference state},
\begin{equation}
\tau_k \;=\; \big(x^{\mathrm{ref}}_k + H_k\,u(x^{\mathrm{ref}}_k,t_k)\big) \;-\; x^{\mathrm{ref}}_{k+1},
\qquad H_k=t_{k+1}-t_k,
\end{equation}
the explicit-Euler step from $x^{\mathrm{ref}}_k$ minus the accurate next reference state. The endpoint
target is the actual cheap-solve error $e = x^{\mathrm{Euler}}_T - x^{\mathrm{ref}}_T$ (same sign as
$\tau_k$, so a faithful reconstruction has positive cosine).

\paragraph{Tangent propagation.} $\Phi(T,t_{k+1})\,\tau_k$ is the directional derivative of the near-exact
endpoint map $S_{t_{k+1}\to T}$ (Heun to data) along $\tau_k$, estimated by a central finite difference at
the reference state,
\begin{equation}
c_k \;=\; \frac{S_{t_{k+1}\to T}(x^{\mathrm{ref}}_{k+1}+\varepsilon\tau_k)
- S_{t_{k+1}\to T}(x^{\mathrm{ref}}_{k+1}-\varepsilon\tau_k)}{2\varepsilon}, \qquad \varepsilon=0.05,
\end{equation}
in the verified linear-response regime (the response is linear in $\varepsilon$ across a factor of four;
\S\ref{sec:transport}). This re-linearizes along the reference path at each $k$ rather than composing one
global Jacobian. The full-state reconstruction is $\hat e=\sum_k c_k$; the no-propagation baseline is
$\hat e_0=\sum_k\tau_k$ (residuals summed without transport). We report the vector cosines $\cos(\hat e,e)$
and $\cos(\hat e_0,e)$ and the relative error $\lVert\hat e-e\rVert/\lVert e\rVert$, reconstructing the
vector and only then taking norms, never norm-then-sum. The propagated reconstruction reaches median cosine
$0.81$ (CelebA, $n{=}32$; bootstrap $95\%$ CI $[0.75,0.85]$) and $0.87$ (ImageNet, $n{=}32$; $[0.82,0.91]$),
against $0.71$ and $0.77$ without propagation, at norm ratio $\lVert\hat e\rVert/\lVert e\rVert = 0.98$ and
$1.03$ (overall magnitude recovered, not only direction) and median relative error $0.65$ and $0.50$.

\paragraph{Regional decomposition.} With $P_i$ the projection onto region $i$, the source-$j$ contribution
propagates the region-restricted residual $P_j\tau_k$: the diagonal $g_i=\sum_k P_i\,c_k^{(i)}$ (where
$c_k^{(i)}$ is the central difference along $P_i\tau_k$, i.e.\ region $i$'s own injection read back at its
source) and the transported-in part $r_i=P_i\hat e - g_i$. Against the actual per-region error
$\lVert P_i e\rVert^2$ we compare, in increasing information, $\FC_i$, the local signed residual
$\lVert\sum_k P_i\tau_k\rVert^2$, the diagonal $\lVert g_i\rVert^2$, and the full $\lVert P_i\hat e\rVert^2$,
and report the partial rank correlation $\rho\big(r_i,\ \lVert P_i e\rVert^2 \mid \text{diagonal}\big)$ over
$64$ strided source regions. The within-image median $\rho$ against $\lVert P_i e\rVert^2$ rises with the
information used: $\FC_i$ $0.66/0.55$, unsigned residual $\sum_k\lVert P_i\tau_k\rVert$ $0.69/0.58$, signed
residual $0.63/0.56$, diagonal $0.42/0.33$, and full $0.80/0.82$ (CelebA\,$n{=}32$ / ImageNet\,$n{=}24$;
image-level bootstrap $95\%$ CI on the full term $[0.74,0.85]$ and $[0.74,0.87]$). A
region's own injection propagated back to itself is the weakest of these, and only all-source transport
closes the reconstruction. Because $r_i$ is built from the true residuals and the model's Jacobian, the
non-circular content is the \emph{contrast} between the diagonal ($\rho\approx0.33$--$0.42$) and off-diagonal
(partial $\rho\approx0.66$--$0.69$, CIs $[0.61,0.79]$ and $[0.63,0.77]$) terms, not the full reconstruction's association with $e$.

\paragraph{Source-count and structure controls.} To rule out that the off-diagonal simply beats the
diagonal by aggregating more terms, we accumulate each source's full propagated response $C_j$ and
reconstruct region $i$'s error from subsets of sources ($n{=}16$ per model). The within-image median
$\rho$ climbs with the number of off-diagonal sources included, from the diagonal alone
($\rho{=}0.32$/$0.30$, CelebA/ImageNet) to all $63$ ($0.74$/$0.56$), so the reconstruction is carried by
many sources, not a few. The decisive control is a sign-randomizing null: it keeps every block norm $\lVert P_i C_j\rVert$
\emph{exactly}, and hence the incoherent budget $\sum_j\lVert P_i C_j\rVert^2$ at each destination, and
randomizes only the relative signs, destroying the signed coherence of the superposition. It reconstructs
error at $\rho{=}0.39$ $[0.22,0.46]$ (CelebA) and $0.27$ $[0.17,0.43]$ (ImageNet), comparable to the diagonal
($0.32$/$0.30$) and well below the correctly-signed transport ($0.74$/$0.56$). So the correct signed
superposition of the propagated residuals carries information beyond the magnitudes of the incoming
contributions; the reconstruction cannot be reproduced from the incoherent budget alone. The null isolates
signed coherence rather than the learned Jacobian as such, since each $C_j$ combines the true residual with
its propagation. An unconstrained matched-\emph{global}-norm
random field gives $\rho{\approx}0$ ($-0.02$/$-0.05$), but it spreads energy nearly uniformly across
destinations and therefore does not preserve the incoming-energy budget; we report the sign-preserving null
as the honest control.

\paragraph{Operator-null family (protocol frozen before evaluation).} Three further nulls probe the
structure of the propagated contributions, with the protocol and decision rule frozen before the runs
($n{=}16$ per model, the source-count protocol above; this rerun reproduces the committed medians).
Because each $C_j$ combines residual content with its propagation, these nulls test the structure of the
propagated contributions, not the learned Jacobian in isolation. (1)~\emph{Destination permutation}: each
source $j$'s per-destination blocks $\{P_i C_j\}$ are randomly reassigned among the $64$ evaluated
destinations, preserving every block's content, norm, and sign and destroying only the
source$\to$destination correspondence. The reconstruction collapses to $\rho = 0.004$ $[-0.011, 0.018]$
(CelebA) and $0.005$ $[-0.006, 0.020]$ (ImageNet), below even the sign null, so the destination
correspondence of the propagated contributions is the single most load-bearing ingredient.
(2)~\emph{Time-shuffled propagation}: each step's residual is
injected at a deranged step slot on the same trajectory; $\rho$ falls to $0.59$ $[0.36, 0.69]$ on CelebA
and $0.26$ $[0.15, 0.33]$ on ImageNet. (3)~\emph{Cross-image dynamics}: one image's residuals propagated
through another image's trajectory, scored against the first image's error ($15$ pairs per model);
$\rho = 0.59$ $[0.36, 0.73]$ and $0.49$ $[0.24, 0.59]$. Under the frozen rule (the full reconstruction's
median must exceed every
null's upper CI on both models), correspondence, sign, and timing specificity hold on both models (timing
narrowly on CelebA), while cross-image dynamics remains within CI of the full reconstruction on ImageNet,
so we do not claim image specificity. Because every null is measured on the same images as the full
reconstruction, paired differences are the sharper statistic (median per-image
$\Delta\rho = \rho_{\mathrm{full}} - \rho_{\mathrm{null}}$, image-level bootstrap; for cross-image
dynamics the prediction in pair $i$ concerns image $i{-}1$'s error, so it is paired with image
$i{-}1$'s full reconstruction): destination permutation $+0.72$ $[0.61, 0.83]$ / $+0.54$ $[0.47, 0.69]$
(CelebA/ImageNet), sign $+0.27$ $[0.20, 0.44]$ / $+0.28$ $[0.18, 0.39]$, time shuffle $+0.16$
$[0.10, 0.20]$ / $+0.29$ $[0.12, 0.62]$, cross-image $+0.18$ $[0.08, 0.29]$ / $+0.12$ $[-0.01, 0.23]$.
The paired analysis agrees with the frozen rule and adds one nuance: image specificity is supported on
CelebA but not on ImageNet, so we do not claim it across models. Wrong-image transport retains
substantial signal on both models, and wrong-time transport retains it on CelebA while falling to the
diagonal's level on ImageNet, so how much of the operator's spatial structure is stationary and shared
across images is itself model-dependent. Artifacts:
\texttt{results/recon/opnulls\_\{celeba,imagenet\}.json}.

\paragraph{Region-scale robustness (fixed source budget).} We recompute the decomposition on CelebA at
three region grids ($32{\times}32$, $16{\times}16$, $8{\times}8$), in each case decomposing over the same
$64$ strided sources. At every scale the full reconstruction beats the diagonal (full $\rho$
$0.26$/$0.74$/$0.87$, diagonal $0.12$/$0.32$/$0.68$; vector cosine $0.51$/$0.66$/$0.74$), so signed
all-source transport adds over own injection at every grid (the sign-preserving null above is measured at
the $16{\times}16$ grid). The absolute correlations rise from fine to coarse
grids, but this is confounded and we do not read it as scale-invariance: a fixed $64$-source budget covers
$6\%$, $25\%$, and $100\%$ of regions at the three scales, and finer partitions also make each region's error
noisier. The scale-independent claim is only the ordering (full $>$ diagonal $>$ null); the main text uses the
$16{\times}16$ grid.

\paragraph{Robustness across step size.} The reconstruction is not tuned to one budget. Sweeping the cheap
NFE at a fixed near-exact reference (CelebA, $n{=}8$), the transported reconstruction holds
($\cos = 0.83, 0.78, 0.76, 0.76, 0.79$ at NFE $4, 6, 8, 10, 16$) while the no-propagation baseline degrades
monotonically ($0.81, 0.72, 0.65, 0.61, 0.53$), so the transport increment grows from $+0.02$ to
$+0.26$ as the step shrinks and becomes the dominant structure at finer stepping rather than a
coarse-NFE artifact. The residual ${\sim}0.2$ shortfall from unity may arise from nonlinear defect interactions, the
finite-difference and reference-solver approximations, and evaluation along the reference rather than the perturbed trajectory; we do not decompose it here.

\subsection{The measured error-determining window $W^*$}
Where should $\FC$ be read? We measure, per time-bin, the within-image $\rho$ between
that bin's acceleration and the final per-region error, early$\to$late: SiT (flow,
latent) \textbf{0.55}, 0.29, 0.18, 0.05, 0.05 ($W^*$ early); ADM (diffusion, pixel) 0.31,
\textbf{0.60}, \textbf{0.58}, 0.47, 0.28 ($W^*$ mid); DiT (diffusion, latent) 0.22, 0.22,
\textbf{0.48}, \textbf{0.51}, 0.23 ($W^*$ mid/late). The dominant bins define the
\emph{error-determining window} $W^*$, and $W^*$ coincides with the prediction horizon
measured in \S\ref{sec:horizon}: the horizon and the error-determining epoch are the same
measurement. Difficulty is \emph{not} a fixed spatial template revealed early
(early$\leftrightarrow$late acceleration $\rho$ = 0.10 / $-0.14$ / 0.47); each model has
an error-determining epoch, and $\FC$ read there predicts the error.

\section{Experimental details}
\label{app:details}

\subsection{Models}
\begin{itemize}[leftmargin=1.6em,itemsep=1pt,topsep=2pt]
\item \textbf{External 256px models (analysis, \S\ref{sec:landscape}--\S\ref{sec:horizon};
  we train nothing at this scale).}
  \textbf{SiT-XL/2} \citep{ma2024sit}: 675M, flow (linear interpolant), latent (SD-VAE),
  ImageNet-256, public checkpoint; analysis at cfg $=1$, $n=256$ images (stable vs.\
  $n=128$).
  \textbf{DiT-XL/2} \citep{peebles2023scalable}: diffusion, latent, ImageNet-256, public
  checkpoint, DDIM, cfg $=1$, $n=256$; both $x_0$- and $\varepsilon$-analog drift signals
  computed, and their window profiles agree, so the reported result is analog-independent.
  \textbf{ADM} \citep{dhariwal2021diffusion}: diffusion, \emph{pixel-space} U-Net,
  ImageNet-256, public checkpoint, DDIM, $\varepsilon$-analog, $n=128$.
\item \textbf{From-scratch 256px models (intervention, \S\ref{sec:results}).}
  Class-conditional \textbf{70M FM-DiT} on ImageNet-256 and unconditional \textbf{33M
  FM-DiT} on CelebA-HQ-256, both latent-space (SD-VAE), trained 100k steps in paired arms
  (baseline vs.\ \methodshort{}) with identical architecture, data order, noise,
  timesteps, and conditioning masks, at matched optimization steps and data exposure
  (\methodshort{} adds one forward per step, measured $1.9\times$ step time, so this is not matched
  training compute).
\item \textbf{Controlled complements ($32\times32$).} Compact FM-DiT models trained from
  scratch (9.8M, 32M, and 77M capacity variants and an independent seed), used only for effects
  that intrinsically require retraining (cross-seed stability, emergence,
  cross-capacity, \S\ref{sec:stable}--\S\ref{sec:learned}) and for the controlled
  studies (selector comparison, $\lambda$-sweep, mechanism diagnostic).
\end{itemize}

\subsection{Reference discrepancy and signal estimation}
\textbf{Per-region integration error:} per-region MSE between a cheap solve (few-step
Euler/DDIM, e.g.\ NFE 6) and a near-exact reference solve (Heun-200) of the same ODE from
the same noise. This target is measured independently of the signal: it compares final
states only and knows nothing of the predictor.
\textbf{Regions:} $16\times16$\,px tokens at $256^2$ resolution (latent models:
token-aligned latent patches).
\textbf{Time convention and windows:} $t$ runs $1\to0$ from noise to data; the
observation cutoff $f$ of \S\ref{sec:horizon} is the fraction of sampling elapsed from
noise. ``First ${\sim}10\%$'' $= t\in[1.0,0.9]$; ``first 30\%'' $= t\in[1.0,0.7]$.
\textbf{Rank statistics:} all correlations are within-image Spearman $\rho$ (per-bin
$W^*$ profiles likewise), invariant to the overall $\dt$ scale of
Definition~\ref{def:fc}.
\textbf{Estimator note:} Definition~\ref{def:fc} is $\sum\lVert\Delta u\rVert$; the
implementation uses estimators that are rank-equivalent within a fixed uniform window
(mean $\lVert\Delta u\rVert$; time-weighted mean of $\lVert\Delta u\rVert/\dt$),
differing from $\sum\lVert\Delta u\rVert$ only by positive constants that cancel under
within-image ranking.
\textbf{Trajectory used for \signalshort{}:} unless stated otherwise the window $W$ is
sampled along the recorded reference trajectory. To test what survives without it, we
recomputed \signalshort{} from the cheap NFE-6 Euler solve itself on SiT with a held-out
seed ($n=96$, disjoint from every other SiT analysis). Within-image $\rho$ against the
final per-region error is $+0.26$ using all six velocity evaluations and
$+0.32/+0.30/+0.29/+0.27$ using the first $2/3/4/5$, versus $+0.47$ (full-window) and
$+0.58$ (early-window) for reference-trajectory \signalshort{} on the same images. The
correlation falls under coarse discretization but does not collapse. Among the prefixes
evaluated on this held-out set the earliest ($k=2$) was the highest. That observation was
made after seeing the results, so we froze $k=2$ and the metrics in advance and ran a
second held-out seed as a confirmatory test. On the confirmatory seed the frozen primary
metric, top-15\% error capture by the two-evaluation cheap prefix, is $0.245$
$[0.23, 0.26]$ (image-level bootstrap) against an oracle ceiling of $0.352$ and a random
baseline of $38/256=0.148$, with mean within-image $\rho = +0.40$ $[+0.36, +0.44]$, median
$+0.42$, and $3\%$ of images at $\rho \le 0$. Comparators behave as on the exploratory
seed (reference early-window $\rho = +0.62$, capture $0.29$; cheap full $\rho = +0.30$,
capture $0.21$). The two seeds are reported separately, never pooled. A cross-model
replicate on our own CelebA-HQ-256 flow model under the same frozen protocol (fresh seed,
$n=96$, unconditional) gives capture $0.262$ $[0.25, 0.28]$ against an oracle of $0.408$,
with $\rho = +0.46$ $[+0.43, +0.50]$ and $1\%$ of images at $\rho \le 0$; the frozen $k=2$
prefix was again the strongest among the prefixes, without reselection. Per-image
statistics and provenance are committed with the code (\texttt{results/sit/},
\texttt{results/celebahq/}).

\subsection{Matched-budget selector comparison (controlled study)}
\label{app:selector}
The comparison referenced in \S\ref{sec:notlandscape}: on the controlled $32\times32$
model, a fixed per-region compute budget (top-10\% of regions selected for extra solver
accuracy at NFE 10) is allocated by different signals, varying \emph{only} the selection
signal at identical budget (50k-sample FID, lower is better). Selecting the highest-\signalshort{}
regions gives \signalshort{} \textbf{14.35} $<$ random 15.35 $\approx$ attention-entropy 15.41 $<$ edge
15.90 $<$ frequency 18.40 $\approx$ velocity-magnitude 18.52: among the reactive signals only \signalshort{}
beats random, and edge, frequency, and velocity-magnitude actively mislead.

\textbf{An important counter-result.} The FID-optimal allocation is the \emph{reverse} of the error-optimal
one. Refining the \emph{lowest}-\signalshort{} (smooth, easy) regions instead reaches FID $\mathbf{12.40}$,
better than refining the highest-\signalshort{} regions ($14.35$) and than random ($15.35$), and the same
low-\signalshort{} choice also wins on perceptual LPIPS-to-reference ($0.0286$ vs $0.0299$). Yet by pixel
fidelity to the reference solve (the integration error \signalshort{} is meant to track), the
high-\signalshort{} choice is best (MSE $0.0351$). So \signalshort{} does identify where integration error
concentrates, but reducing that error is not the same as improving perceptual sample quality: smooth regions
matter more to FID than their small integration error suggests. This dissociation is exactly why we treat
\signalshort{} as a diagnostic of integration difficulty and do \emph{not} propose it as a sampling policy
(\S\ref{sec:limitations}). Caveat: \signalshort{} also selects spatially coherent regions, so part of the
margin may be a coherence effect.

\subsection{Intervention training details}
\begin{figure}[tbp]
\centering
\includegraphics[width=0.86\linewidth]{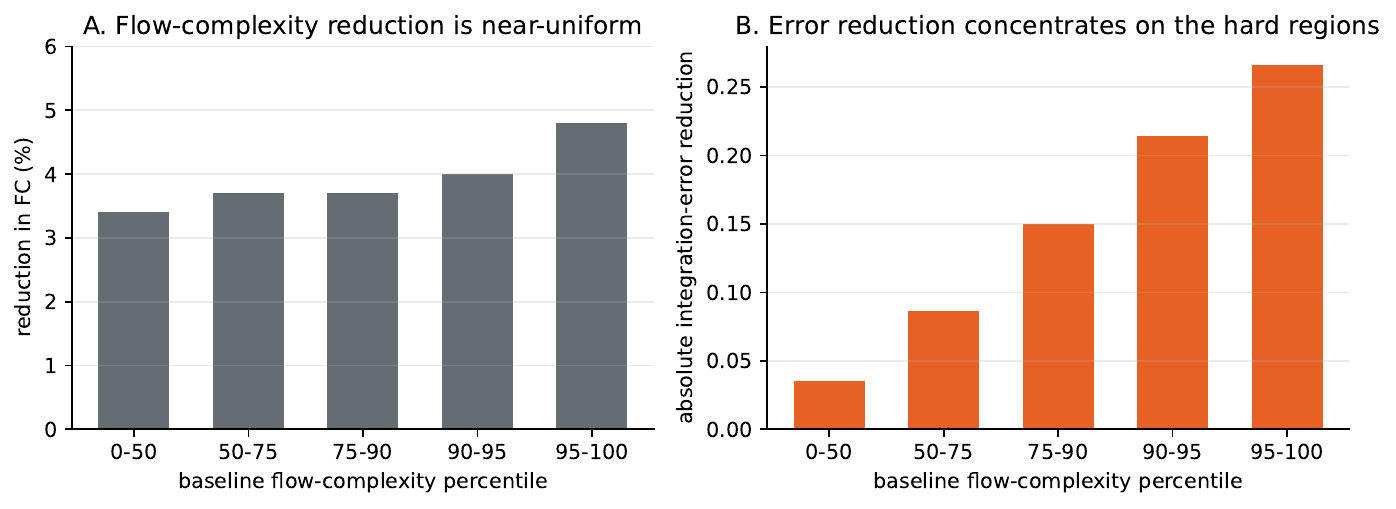}
\caption{\textbf{The penalty's benefit lands on the difficult landscape.} Regions binned by baseline
\signalshort{} percentile (CelebA, $n{=}32768$ regions, $\lambda{=}0.3$ vs.\ baseline at matched $100$k
steps). \textbf{(A)} the penalty lowers \signalshort{} by a near-uniform $3.4$--$4.8\%$ across every
difficulty bin. \textbf{(B)} the resulting absolute integration-error reduction climbs about $7.5\times$
from the easy half to the hardest $5\%$, concentrating the improvement on the high-\signalshort{} regions
\S\ref{sec:landscape} flags as difficult. Average \signalshort{} is thus the wrong summary of the effect.}
\label{fig:mechanism}
\end{figure}

Objective as in Eq.~\ref{eq:fcm}: $\bar v$ is the detached current velocity;
$\dt_{\mathrm{eff}}$ clips the lookahead at the per-sample time boundary; for
class-conditional training both evaluations share one classifier-free-guidance drop mask
(so the penalty measures trajectory change, not conditioning change); $\lambda = 0.3$
throughout, selected by the controlled $\lambda$-sweep below. Cost: one additional
forward evaluation per training step (measured $1.9\times$ step time at equal batch, $0.169$s vs.\
$0.087$s; no
additional parameters, no teacher, no reflow stage). FID protocol: 50k generated samples
against standard reference statistics for each dataset; ImageNet-256 additionally
evaluated at cfg $=1.5$ (\S\ref{sec:results}). The $32\times32$ ablations predate the
shared-conditioning-mask refinement that all 256px runs include
(\S\ref{sec:limitations}).
\begin{table}[h]
\centering\small
\setlength{\tabcolsep}{4pt}
\resizebox{\linewidth}{!}{%
\begin{tabular}{lccccc}
\toprule
model & data ($C{\times}$res) & arch (hid/dep/heads, p) & params & batch & steps \\
\midrule
CIFAR-10 (ablation)   & $3{\times}32$ px               & $384/12/6$, p2 & 32.6M & 128 & 400k \\
CelebA-HQ-256         & $4{\times}32$ latent           & $384/12/6$, p2 & 33M   & 128 & 100k \\
LSUN-Church-256       & $4{\times}32$ latent           & $384/12/6$, p2 & 33M   & 128 & 100k \\
ImageNet-256          & $4{\times}32$ latent, 1000 cls & $512/16/8$, p2 & 70M   & 64  & 100k / 400k \\
\bottomrule
\end{tabular}}
\caption{Training configurations. All arms use AdamW, a constant learning rate $10^{-4}$ (no warmup or
decay schedule), weight decay $0$, gradient clipping $1.0$, EMA $0.9995$ (CIFAR $0.9999$), and mixed
precision. \methodshort{} adds a
per-step penalty ($\lambda=0.3$) with a finite-difference lookahead $\dt=0.05$ and one extra forward per step;
paired arms share every other setting and the random seed.}
\label{tab:hparams}
\end{table}

\subsection{$\lambda$-sweep ($32\times32$, controlled)}
\label{app:lambda}
$\lambda = 0.3$ improves every few-step budget with no high-NFE cost, and $\lambda = 1.0$
over-straightens (gains at low NFE, a regression at NFE 16). In the mechanism diagnostic (binned
$\Delta$\signalshort{} vs.\ $\Delta$error), the \signalshort{} reduction is near-uniform
across difficulty percentiles but the error reduction is strongly tail-weighted, so average
\signalshort{} is the wrong statistic for the mechanism. That is why we report the per-percentile
analysis of \S\ref{sec:intervention}.
\begin{table}[h]
\centering
\small
\caption{$\lambda$-sweep on CIFAR-10 ($32\times32$), FID vs.\ NFE, $10$k samples, matched
$100$k training steps, same noise. Lower is better; bold marks the best in each row.
$\lambda=0.3$ wins or ties at every budget through NFE\,12 and carries no high-NFE penalty;
$\lambda=1.0$ edges it at NFE\,6 but regresses past baseline by NFE\,16 (over-straightening);
$\lambda=0.5$ is dominated. At $50$k samples the $\lambda=0.3$ gain holds (NFE\,6: $31.8\to29.2$).}
\label{tab:lambda}
\begin{tabular}{lcccc}
\toprule
NFE & baseline & $\lambda=0.3$ & $\lambda=0.5$ & $\lambda=1.0$ \\
\midrule
6  & 33.9 & 31.5 & 32.1 & \textbf{31.4} \\
8  & 28.7 & \textbf{27.1} & 27.8 & 27.7 \\
10 & 26.6 & \textbf{25.2} & 26.2 & 26.7 \\
12 & 24.5 & \textbf{23.5} & 24.6 & 25.0 \\
16 & \textbf{23.1} & 23.3 & 24.7 & 25.2 \\
\bottomrule
\end{tabular}
\end{table}

\subsection{Training-run variance (controlled model)}
\label{app:seedvar}
Three independently seeded paired runs on the $32\times32$ model (baseline vs.\ \methodshort{}
$\lambda{=}0.3$, matched seed and data order, $50$k steps each; $10$k-sample FID). The paired
$\Delta$FID $=$ baseline $-$ \methodshort{} (positive $\Rightarrow$ \methodshort{} better):
\begin{table}[h]
\centering\small
\setlength{\tabcolsep}{5pt}
\begin{tabular}{lccccc}
\toprule
NFE & 6 & 8 & 10 & 12 & 16 \\
\midrule
seed 0 & $+5.07$ & $+0.61$ & $-1.84$ & $-3.27$ & $-4.17$ \\
seed 1 & $+5.10$ & $+2.73$ & $+1.24$ & $+0.48$ & $+0.22$ \\
seed 2 & $+10.73$ & $+5.28$ & $+3.09$ & $+1.49$ & $+0.89$ \\
\midrule
mean $\pm$ std & $\mathbf{+6.97{\pm}3.26}$ & $\mathbf{+2.87{\pm}2.34}$ & $+0.83{\pm}2.49$ & $-0.43{\pm}2.50$ & $-1.02{\pm}2.75$ \\
\bottomrule
\end{tabular}
\end{table}
\\
The few-step improvement (NFE $6$--$8$) is consistent: all three seeds favor \methodshort{} and the
mean is well outside the seed spread. By NFE $10$--$16$ the paired delta is within seed noise, so on the
controlled model the gain is a few-step effect, matching the $\lambda$-sweep (Table~\ref{tab:lambda}),
where $\lambda{=}0.3$'s advantage also concentrates at low NFE. (These are $50$k-step models; the headline
$256$px results are at $100$k steps and $50$k-sample FID.)

\subsection{E4: training-progress and compute-matched frontier}
\label{app:e4}
\begin{table}[h]
\centering\small
\caption{ImageNet-256, class-conditional 70M FM-DiT trained from scratch (100k steps,
paired arms at matched optimization steps and data exposure; see the \S\ref{sec:results} protocol note on
per-step compute). FID (50k samples, cfg $=1.0$) by sampling budget (NFE); lower is better.}
\label{tab:imagenet}
\begin{tabular}{lccccc}
\toprule
NFE & 6 & 8 & 10 & 12 & 16 \\
\midrule
baseline & 131.20 & 119.69 & 113.68 & 109.05 & 105.12 \\
\textbf{\methodshort{}} & \textbf{121.65} & \textbf{112.52} & \textbf{107.71} & \textbf{104.28} & \textbf{101.31} \\
$\Delta$ (rel.) & $-7.3\%$ & $-6.0\%$ & $-5.3\%$ & $-4.4\%$ & $-3.6\%$ \\
\bottomrule
\end{tabular}
\end{table}
Paired ImageNet-256 arms (baseline $\lambda{=}0$ vs.\ \methodshort{} $\lambda{=}0.3$), evaluated at the
$\{50,100,200,300,400\}$k checkpoint ladder with $10$k-sample FID (lower is better). At matched optimization
steps \methodshort{} leads at every checkpoint and the NFE-6 advantage widens as the baseline strengthens
(Table~\ref{tab:e4}), so the few-step gain is not an undertraining artifact.
\begin{table}[h]
\centering\small
\caption{E4 NFE-6 FID ($10$k samples) across the training ladder. $\Delta$ is the relative \methodshort{}
advantage $(\text{baseline}-\text{\methodshort{}})/\text{baseline}$.}
\label{tab:e4}
\begin{tabular}{lccccc}
\toprule
steps & 50k & 100k & 200k & 300k & 400k \\
\midrule
baseline & 149.1 & 134.1 & 118.7 & 113.0 & 108.6 \\
\textbf{\methodshort{}} & \textbf{142.5} & \textbf{124.6} & \textbf{108.9} & \textbf{100.8} & \textbf{96.3} \\
$\Delta$ & $+4.4\%$ & $+7.1\%$ & $+8.3\%$ & $+10.7\%$ & $+11.4\%$ \\
\bottomrule
\end{tabular}
\end{table}
The compute-matched picture differs. Because a \methodshort{} step costs a measured $1.9\times$ a baseline
step, we place both arms on a training-compute axis (steps $\times$ per-step cost) and read the baseline curve
at \methodshort{}'s compute. At equal compute the baseline reaches \emph{lower} few-step FID than
\methodshort{} across the reachable range: at NFE $6$ the baseline leads by $4$--$7$ FID up to
$\sim\!200$k baseline-equivalent compute and the two draw level only near $\sim\!380$k (\methodshort{}@$200$k
$108.9$ vs baseline@$400$k $108.6$), and at NFE $10$ and $16$ the baseline leads throughout. So on ImageNet
the extra forward per step is not recovered in few-step FID, the reverse of the CelebA result
(\S\ref{sec:results}). Per the advance-specified decision rule this keeps \methodshort{} as a mechanism and a
faster-per-step-convergence result on ImageNet, not a compute-matched sampler advantage. A single training
run per arm here; the small-model replication (Appendix~\ref{app:seedvar}) provides supporting evidence but
does not estimate ImageNet training-run variance.

\subsection{Region-scale robustness (SiT)}
\label{app:multiscale}
The two landscape conclusions --- moderate concentration and \signalshort{} predictivity --- do not depend on
the region grid. On SiT-XL/2 (ImageNet-256, $n{=}96$) we sweep the region size:
\begin{table}[h]
\centering\small
\begin{tabular}{lcccc}
\toprule
grid & regions & FC--error $\rho$ & top-$10\%$ share & Gini \\
\midrule
$32{\times}32$ & $1024$ & $0.54$ & $31\%$ & $0.44$ \\
$16{\times}16$ & $256$  & $0.57$ & $27\%$ & $0.35$ \\
$8{\times}8$   & $64$   & $0.61$ & $22\%$ & $0.29$ \\
\bottomrule
\end{tabular}
\caption{Region-scale sweep on SiT. FC--error $\rho$ is stable ($0.54$--$0.61$) and concentration stays
moderate (top-decile $22$--$31\%$, Gini $0.29$--$0.44$) at every grid; finer partitions are slightly more
concentrated, as expected when smaller regions resolve more heterogeneity. The main text uses $16{\times}16$.}
\label{tab:multiscale}
\end{table}

\subsection{Numerical variation baselines}
\label{app:numbase}
Within-image Spearman $\rho$ (mean over images; image-level bootstrap $95\%$ CIs in the committed
artifacts) between each variation statistic, computed from the reference record over the stated window,
and the final per-region error ($n{=}96$ per backend, fresh seeds: SiT seed 4, CelebA seed 5; all
signals, windows, and backends reported, none selected).
\begin{table}[h]
\centering\small
\setlength{\tabcolsep}{5pt}
\begin{tabular}{l cc cc}
\toprule
 & \multicolumn{2}{c}{SiT-XL/2} & \multicolumn{2}{c}{CelebA-HQ} \\
signal & full & early & full & early \\
\midrule
\signalshort{} (mean $\lVert\Delta u\rVert$) & 0.53 & 0.65 & 0.63 & 0.73 \\
signed cumulative $\lVert\sum\Delta u\rVert$ & 0.43 & \textbf{0.73} & 0.47 & 0.75 \\
squared variation $\sum\lVert\Delta u\rVert^2$ & 0.50 & 0.68 & 0.58 & \textbf{0.77} \\
max step & 0.24 & 0.52 & 0.32 & 0.70 \\
velocity magnitude & 0.43 & 0.25 & 0.31 & 0.19 \\
horizon-weighted $\sum\lVert\Delta u\rVert\!\cdot\!\mathrm{rem}$ & \textbf{0.66} & 0.68 & \textbf{0.75} & 0.76 \\
\bottomrule
\end{tabular}
\end{table}
The variation family clusters well above the magnitude control at the early window on both backends.
Cancellation-aware and horizon-weighted members predict as well as or somewhat better than
\signalshort{}, consistent with the transport account (\S\ref{sec:transport}): the signed cumulative
change is the unpropagated signed-residual accumulation of the reconstruction, and the horizon weight
approximates how much trajectory remains to transport an injection. Artifacts:
\texttt{results/sit/numeric\_baselines\_sit.json},
\texttt{results/celebahq/numeric\_baselines\_celeba.json}.

\subsection{Energy split of the regional transport response}
\label{app:energysplit}
$R_{\mathrm{self}}$ in \S\ref{sec:transport} is a share of summed per-region response norms, which
weights many small contributions heavily. Because the region projections are disjoint, per-region
response energies $\lVert P_i\,\delta x\rVert^2$ sum exactly to the total, so the energy share
$R^{(2)}_{\mathrm{self}} = \lVert P_j\,\delta x\rVert^2 / \sum_i \lVert P_i\,\delta x\rVert^2$ is an
orthogonally additive companion, and the participation ratio
$N_{\mathrm{eff}} = (\sum_i \varepsilon_i)^2/\sum_i \varepsilon_i^2$ with
$\varepsilon_i = \lVert P_i\,\delta x\rVert^2$ counts the effective number of destination regions.
On smaller replicates of the \S\ref{sec:transport} protocol (SiT $n{=}16$, CelebA $n{=}16$, ADM
$n{=}8$; $64$ strided sources out of $256$ per image; image-level medians, bootstrap $95\%$ CIs),
near the noise end the source region retains
$0.18$ $[0.12, 0.25]$ (SiT), $0.40$ $[0.33, 0.43]$ (CelebA), and $0.10$ $[0.07, 0.13]$ (ADM) of the
response energy, and the response spreads over $N_{\mathrm{eff}} = 23$ $[15, 28]$, $9$ $[7, 13]$,
and $43$ $[30, 55]$ effective regions. The source carries more weight than the norm share suggests,
but the majority of response energy still lands elsewhere.
Toward the data end both measures localize ($R^{(2)}_{\mathrm{self}}$ $0.98$/$0.96$/$0.66$,
$N_{\mathrm{eff}}\approx 1$--$2$), consistent with the shrinking-horizon account of
\S\ref{sec:transport}. Artifacts: \texttt{results/propagation/energy\_*.json}.

\subsection{Reproducibility}
Code and configs for every result (training, paired-arm orchestration, reference-discrepancy and
horizon/$W^*$ measurement, and FID evaluation) will be released upon publication.

\end{document}